\documentclass[sigconf]{acmart}

\usepackage{booktabs}
\usepackage{multirow}
\usepackage{graphicx}
\usepackage{xcolor}
\usepackage{microtype}
\usepackage{enumitem}
\setlist{nosep,leftmargin=*}
\setlist[enumerate]{topsep=2pt,itemsep=1pt,parsep=0pt,partopsep=0pt}
\setlist[itemize]{topsep=2pt,itemsep=1pt,parsep=0pt,partopsep=0pt}
\setlist[description]{topsep=2pt,itemsep=1pt,parsep=0pt,partopsep=0pt}
\usepackage{amsmath}

\usepackage{amssymb}
\usepackage{subcaption}
\usepackage{adjustbox}
\usepackage{listings}
\usepackage{float}
\copyrightyear{2026}
\acmYear{2026}
\acmConference[KDD '26 Workshop]{5th Workshop on End-End Customer Journey Optimization}{August 9, 2026}{Jeju, Republic of Korea}

\renewcommand\footnotetextcopyrightpermission[1]{}

\begin{document}

\title{CARRE: Counterfactual Action Retrieval and Reason Evaluation for Explainable Churn Prescription}

\author{MinJoo Kim}
\affiliation{
  \department{Department of Industrial Data Engineering}
  \institution{Hanyang University}
  \city{Seoul}
  \country{Republic of Korea}
}
\email{kmj0921@hanyang.ac.kr}

\author{Sangjin Park}
\authornote{Corresponding authors.}
\affiliation{
  \department{Department of Industrial Data Engineering}
  \institution{Hanyang University}
  \city{Seoul}
  \country{Republic of Korea}
}
\email{psj3493@hanyang.ac.kr}

\author{Seung Hwan Cho}
\authornotemark[1]
\affiliation{
  \department{Department of Industrial Data Engineering}
  \institution{Hanyang University}
  \city{Seoul}
  \country{Republic of Korea}
}
\email{shcho95@hanyang.ac.kr}

%% ─────────────────────────────────────────────────────────────────────
\begin{abstract}
Churn models typically identify high-risk customers but do not specify which feasible retention action should be considered or why that action is appropriate. We present \textbf{CARRE} (\textbf{C}ounterfactual \textbf{A}ction \textbf{R}etrieval and \textbf{R}eason \textbf{E}valuation), a three-stage framework that combines retrieval-augmented candidate generation, cost-aware counterfactual scoring, and large language model (LLM) reasoning. CARRE retrieves a predefined catalog of retention actions, estimates model-predicted churn-risk changes under explicit feature transformations, and generates a structured churn reason and a profile-grounded explanation for the selected action. On the IBM Telco Customer Churn dataset, CARRE achieves 79.8\% greater mean model-predicted risk reduction than the plain SHAP baseline and 80.4\% greater reduction than the cost-controlled SHAP+Cost baseline across 313 high-risk test cases; its cost-normalized efficiency is 10.5\% higher than that of plain SHAP. On a 136-case reason-stratified evaluation sample, diagnosis-driven prompt refinement increases weak-label agreement from 79.4\% to 90.4\%, with no auxiliary-plan constraint violations; because the same sample was used for error diagnosis and re-evaluation, the post-refinement result is not an independent estimate of generalization. For 135 explanations generated using the pre-refinement v2 reason outputs, two cross-vendor LLM judges assign mean scores ranging from 4.02 to 5.00 out of 5, although one judge saturates on actionability, and a deterministic audit finds no contradictions among 66 verifiable profile claims. Retrieval ablations show that $k{=}5$ provides the best evaluated compromise between high candidate coverage and downstream reasoning agreement in this dataset. These results illustrate how retrieval, model-based counterfactual scoring, and language generation can be separated and jointly evaluated in a prototype churn-prescription pipeline.
\end{abstract}



% Running header: show only the lead author followed by "et al."
\renewcommand{\shortauthors}{M. Kim et al.}

\maketitle

%% ─────────────────────────────────────────────────────────────────────
\section{Introduction}
%% ─────────────────────────────────────────────────────────────────────

The growth of digital transformation and the platform economy has made subscription-based revenue models ubiquitous across software, telecommunications, media, and financial services. Unlike traditional markets centered on one-time purchases, subscription models allow customers to cancel at any time, making churn management a persistent operational challenge. In the telecommunications industry, annual churn rates are widely reported in the 15--25\% range, and acquiring a new customer is commonly estimated to cost several times more than retaining an existing one \cite{vafeiadis2015comparison}. This economic pressure has spurred extensive machine learning research on churn prediction. A wide range of approaches have been proposed, from logistic regression and decision trees to gradient-boosted ensembles, with profit-driven formulations addressing the operational cost of retention \cite{verbeke2012new}. Boosted variants of these classifiers consistently outperform their non-boosted counterparts on public telecommunications churn benchmarks \cite{vafeiadis2015comparison}.

With predictive accuracy alone, operational deployment remains fundamentally limited: a churn probability score of 0.85 tells a customer retention manager only that a customer is likely to leave, providing no guidance on \emph{why} the customer might churn or \emph{which specific actions} could reduce that risk. This prediction-prescription gap goes beyond a mere technical limitation and translates into real business losses. Repeating uniform retention campaigns without identifying the dominant observable churn-risk pattern yields poor cost efficiency and fails to deliver appropriate interventions to high-risk customers in time. In telecommunications in particular, where thousands of at-risk customers must be handled simultaneously, individualized prescriptions paired with explanatory rationale are essential for frontline agents to make effective decisions.

Existing approaches address these requirements only in isolation. Post-hoc explanation methods such as SHAP (SHapley Additive exPlanations) decompose model predictions into feature contributions, identifying why a customer has a high churn probability \cite{lundberg2017unified}. However, feature importances do not translate directly into actionable recommendations. Knowing that a customer's month-to-month contract is the most negative feature does not reveal whether to offer a contract upgrade, a price discount, or a service bundle. Counterfactual explanation methods identify minimal profile changes that would flip the model's decision, but focus on hypothetical feature transformations rather than business-feasible interventions with defined costs and eligibility constraints \cite{wachter2017counterfactual,karimi2021algorithmic}. LLM-based systems can generate fluent, personalized text, but without grounding in a predefined action space, they tend to produce generic advice or hallucinated interventions that cannot be operationalized.

A further challenge is \emph{explanation alignment}: the property that a system's explanations maintain mutual consistency across churn diagnosis, action recommendation, and frontline delivery. Specifically, a prescription explanation delivered to a customer-facing agent must simultaneously satisfy three criteria. First, it must faithfully reflect the heuristic churn-risk label derived from the customer profile. Second, it must be logically coherent with the recommended action. Third, it must be expressed in language that the frontline agent can act upon immediately. Prior work has addressed each criterion individually. Work on generated-text faithfulness targets the first \cite{maynez2020faithfulness}, counterfactual recourse research addresses actionability \cite{ustun2019actionable}, and LLM-based explanation systems partially achieve coherence. To our knowledge, few prior systems jointly evaluate profile-grounded diagnosis, constrained action selection, and agent-facing explanation within a single churn-prescription pipeline.

This paper proposes \textbf{CARRE}, a unified pipeline connecting prediction, prescription, and explanation through three tightly coupled components:
\begin{enumerate}[leftmargin=1.5em]
  \item \textbf{RAG-based candidate retrieval}: A FAISS index over six business action documents retrieves the top semantically relevant intervention candidates for each customer profile.
  \item \textbf{Counterfactual (CF) optimization}: A calibrated logistic regression model simulates each retrieved candidate as a counterfactual intervention and scores it by jointly weighing the reduction in predicted churn probability against the normalized operational cost of the action. The candidate that best balances these two objectives is selected as the prescription.
  \item \textbf{LLM reasoning with CoT}: An LLM equipped with structured classification criteria and chain-of-thought (CoT) prompting identifies the customer's churn reason and generates a natural-language explanation aligned with the retrieved action context.
\end{enumerate}

By separating semantic search (Stage 1), cost-aware optimization (Stage 2), and natural-language explanation (Stage 3), each component operates within its own strengths. CARRE is built on a core design philosophy in which the LLM's role is \emph{reasoning}, not retrieval or optimization.

The main contributions of this paper are as follows:
\begin{itemize}[leftmargin=1.5em]
  \item An end-to-end RAG+CF+LLM framework for personalized churn prescription, evaluated through a multi-layer protocol combining CF metrics with bootstrap confidence intervals, weak-label agreement, a cross-vendor two-judge automated evaluation, and a deterministic grounding check.
  \item A controlled \emph{SHAP+Cost} baseline showing that CARRE retains an 80.4\% model-predicted risk-reduction advantage under identical cost penalization, together with a $\lambda$ sweep revealing a sharp cost-penalty regime boundary rather than a smooth trade-off.
  \item An exploratory four-model comparison revealing substantial model-specific variation and category-specific failure patterns; because model scale, vendor, architecture, and inference settings are confounded, the experiment does not isolate the cause of these differences.
  \item A prompt-development analysis showing that explicit priority rules and action withholding improve instruction-following fidelity, followed by diagnosis-driven v3 refinement that raises agreement from 79.4\% to 90.4\% on the same 136-case reason-stratified sample; we report this as a post-refinement development result rather than an independent generalization estimate.
  \item A two-step Stage~3 design separating reason classification and auxiliary-plan generation (Stage~3-a, \texttt{cf\_action} withheld) from final explanation generation (Stage~3-b, given reason and \texttt{cf\_action}), targeting the explanation--prescription incoherence that arises when the cost-optimal action and the rule-consistent one diverge (exact auxiliary-plan--action match is 3/135, or 2\%, among explained cases).
\end{itemize}

The remainder of this paper reviews the relevant literature, formalizes CARRE and its action space, describes the experimental protocol, reports quantitative and qualitative results, and concludes with limitations and future directions.

%% ─────────────────────────────────────────────────────────────────────
\section{Related Work}
\label{sec:related}
%% ─────────────────────────────────────────────────────────────────────

\subsection{Churn Prediction, Probability Calibration, and Prescriptive Analytics}

A systematic comparison of classifiers---logistic regression, SVM, decision trees, naive Bayes, artificial neural networks, and their boosted variants---on a telecommunications churn dataset showed that boosting consistently improves over the plain classifiers, with an AdaBoost-boosted polynomial-kernel SVM attaining the best accuracy and F-measure \cite{vafeiadis2015comparison}. Purely statistical evaluation has been critiqued for ignoring operational costs, motivating profit-driven frameworks that jointly account for retention cost and churn risk \cite{verbeke2012new}. Since high-performing classifiers often produce scores that are not directly interpretable as probabilities, calibration via isotonic regression or Platt scaling is applied \cite{niculescu2005predicting}; calibrated probabilities are a critical input to CARRE's Stage~2 CF scoring.

The prescriptive analytics literature distinguishes \emph{descriptive}, \emph{predictive}, and \emph{prescriptive} systems \cite{lepenioti2020prescriptive}, and most deployed churn systems remain at the predictive layer. An engagement-based retention recommender was proposed for e-commerce, but without explaining why a recommendation was selected or how it addresses each customer's churn drivers \cite{vanderveld2016engagement}. CARRE closes this gap by completing the full loop from prediction to prescription to explanation.

\subsection{Counterfactual Explanations}

A counterfactual explanation presents the minimal profile change that would alter the current prediction. Counterfactual explanations were formalized as minimal feature transformation, generated without accessing model internals \cite{wachter2017counterfactual}; actionability constraints were then introduced to keep proposed changes feasible \cite{ustun2019actionable}, and \emph{algorithmic recourse} extended this with structural causal models and a difficulty-proportional cost function \cite{karimi2021algorithmic}. Diverse counterfactual sets exposing the Pareto frontier can also be generated \cite{mothilal2020explaining}. CARRE adopts this cost-aware formulation but restricts recourse to a predefined business action vocabulary rather than arbitrary feature transformations, ensuring operational feasibility for every prescription.

\subsection{Retrieval-Augmented Generation and LLM-Based Explanation}

RAG combines a pre-trained LLM with an external retriever, reducing hallucination on knowledge-intensive tasks \cite{lewis2020retrieval}. Dense retrieval improves recall over sparse matching \cite{karpukhin2020dense}, multi-hop variants handle complex questions \cite{xiong2021answering}, and Self-RAG trains the model to decide when to retrieve and to verify its own answers \cite{asai2023self}.

\begin{figure*}[t]
  \centering
  \IfFileExists{figure.png}{%
    \includegraphics[width=0.85\textwidth]{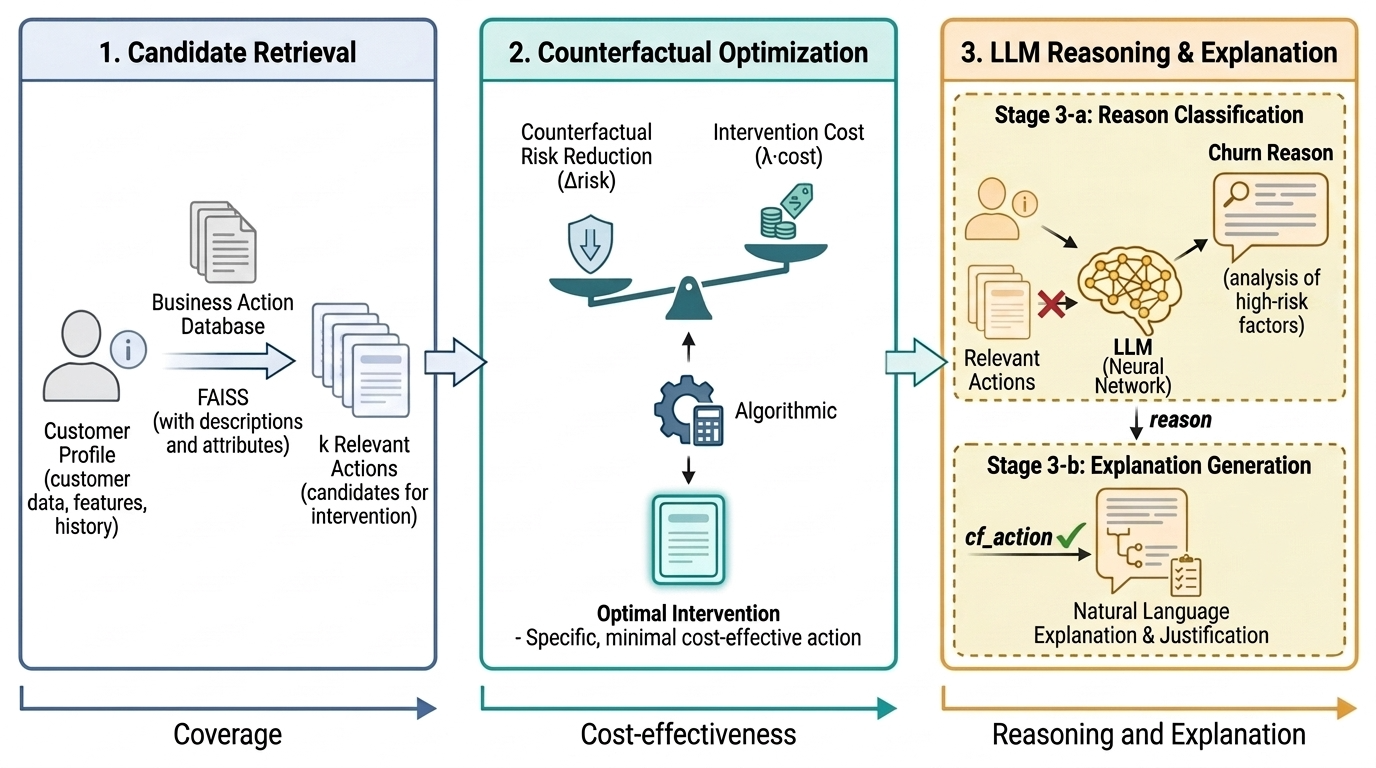}%
  }{%
    \fbox{\parbox[c][4.0cm][c]{0.82\textwidth}{\centering
      Pipeline figure placeholder\\
      Add the original figure file to the project directory.}}%
  }
  \caption{Overview of the CARRE pipeline. Stage 1 (Candidate Retrieval) collects relevant action documents via FAISS semantic search. Stage 2 (CF Optimization) selects the optimal prescription (\texttt{cf\_action}) by balancing risk reduction and intervention cost. Stage 3 (LLM Reasoning \& Explanation) comprises two sub-steps: Stage~3-a classifies the churn reason from the customer profile (\texttt{cf\_action} withheld to prevent reverse-reasoning bias), and Stage~3-b generates the natural-language explanation given the determined reason and the \texttt{cf\_action} from Stage~2.}
  \Description{Three-stage CARRE architecture: semantic retrieval of approved actions, counterfactual risk-cost optimization, and LLM-based reason classification followed by explanation generation.}
  \label{fig:pipeline}
\end{figure*}

In the churn prescription setting, RAG constrains the LLM's plan space to predefined candidate actions to prevent hallucinated interventions, and provides action-specific details (target customer profile, heuristic expected-effect metadata, eligibility conditions) as context to ground the LLM's explanations. The number of retrieved documents $k$ is a key design decision, since too few risk missing the optimal action, while too many introduce irrelevant context that may confuse LLM reasoning.

On the reasoning side, TalkToModel queries ML predictions through dialogue \cite{slack2023explaining}, CoT prompting improves accuracy on tasks requiring structured rule application \cite{wei2022chain}, and explicitly listing judgment criteria improves rule adherence in constrained generation \cite{honovich2022true,white2023prompt}. CARRE's prompt design integrates CoT instructions and explicit classification criteria from these works, motivated by the reverse-reasoning bias discussed in Section~\ref{sec:framework} and revisited in Section~\ref{sec:discussion}.

%% ─────────────────────────────────────────────────────────────────────
\section{Problem Formulation and CARRE Framework}
\label{sec:framework}
%% ─────────────────────────────────────────────────────────────────────

CARRE is a three-stage pipeline that simultaneously outputs \emph{why} a high-risk customer is likely to churn, \emph{what action} may reduce that risk, and \emph{why that action} is appropriate. As illustrated in Figure~\ref{fig:pipeline}, a customer profile is first transformed into a query to retrieve $k$ semantically similar action documents from a FAISS index (Stage 1). Each retrieved action is then CF-simulated to select the optimal action by balancing risk reduction and cost (Stage 2). Finally, the LLM classifies the churn reason from customer features and generates a natural-language explanation grounded in the retrieved action context (Stage 3). The core design principle is \emph{separation of concerns}. Retrieval handles candidate coverage, optimization handles cost efficiency, and the LLM handles only reasoning and language generation, enabling each stage to be independently evaluated and improved.

\subsection{Action Space and Reason Taxonomy}

CARRE's action space $\mathcal{A}$ (target of prescription) and reason taxonomy $\mathcal{R}$ (target of classification) are grounded in prior work that empirically identifies the recurring drivers of telecommunications churn and candidate retention strategies. Price sensitivity, contract type, payment method, and underutilization of add-on services have been consistently identified as primary churn factors \cite{verbeke2012new}. Price discounts, contract switching, auto-pay enrollment, and technical support or security service additions are correspondingly common retention levers in industry practice, motivating their inclusion in $\mathcal{A}$. Table~\ref{tab:actions} lists the six actions in $\mathcal{A}$ with their descriptions and relative costs, where cost denotes a researcher-defined relative weight capturing the operational and financial difficulty of each intervention, ranging from \texttt{A\_PAYMENT\_AUTOPAY} (1.0, lowest) to \texttt{A\_CONTRACT\_24M} (3.5, highest).

\begin{table}[t]
\centering
\caption{Predefined action space $\mathcal{A}$ with researcher-defined relative costs.}
\label{tab:actions}
\small
\begin{adjustbox}{max width=\linewidth}
\begin{tabular}{llc}
\toprule
Action ID & Description & Cost \\
\midrule
\texttt{A\_PRICE\_DISCOUNT}   & Monthly charge discount campaign   & 3.0 \\
\texttt{A\_CONTRACT\_12M}     & Switch to 12-month contract         & 2.0 \\
\texttt{A\_CONTRACT\_24M}     & Switch to 24-month contract         & 3.5 \\
\texttt{A\_PAYMENT\_AUTOPAY}  & Switch to automatic bank transfer  & 1.0 \\
\texttt{A\_ADD\_ONLINESEC}    & Add online security service         & 2.0 \\
\texttt{A\_ADD\_TECHSUPPORT}  & Add technical support service       & 2.0 \\
\bottomrule
\end{tabular}
\end{adjustbox}
\end{table}

The six reason labels in $\mathcal{R}$ are derived from domain rules applied in priority order. Customers with \texttt{MonthlyCharges} $\geq 80$---a threshold corresponding to the upper quartile of the charge distribution in this dataset, indicating elevated financial exposure---on a month-to-month contract are classified as \texttt{PRICE\_SENSITIVE} (Step 1); those who do not meet Step 1 but hold a month-to-month contract with \texttt{tenure} $\leq 6$ months are classified as \texttt{CONTRACT\_RISK} (Step 2). Customers whose \texttt{PaymentMethod} is electronic check are classified as \texttt{PAYMENT\_FRICTION} (Step 3); those using an internet service but subscribed to neither \texttt{TechSupport} nor \texttt{OnlineSecurity} are classified as \texttt{SUPPORT\_DEFICIT} (Step 4). Customers with \texttt{tenure} $\leq 3$ months are classified as \texttt{LOW\_ENGAGEMENT} (Step 5), and those satisfying none of the above conditions are classified as \texttt{OTHER} (Step 6).

The observable profile pattern represented by each heuristic label is interpreted as follows. \texttt{PRICE\_SENSITIVE} marks the combination of high monthly charges and a flexible month-to-month contract as a study-specific proxy for potential price sensitivity. \texttt{CONTRACT\_RISK} marks short-tenure customers on month-to-month contracts as a proxy for limited contractual commitment. \texttt{PAYMENT\_FRICTION} marks electronic-check use as a study-specific proxy for potential payment-process friction; the label does not establish that the payment method causally increases churn. \texttt{SUPPORT\_DEFICIT} marks internet-service customers without technical support or online security as a proxy for limited support coverage. \texttt{LOW\_ENGAGEMENT} marks customers within their first three months as a proxy for limited service experience. \texttt{OTHER} covers profiles not captured by Steps 1--5.

These rules serve as \emph{weak labels} for LLM agreement evaluation. The thresholds, labels, and priority order are study-specific weak-label heuristics designed to create a reproducible instruction-following task; they are not expert-validated causal diagnoses of why a customer will churn. The priority order is a researcher-defined decision rule used to make the weak-label task deterministic when multiple conditions hold.

\subsection{Problem Formulation}

Let $\mathbf{x} \in \mathbb{R}^d$ be a customer feature vector and $f: \mathbb{R}^d \to [0,1]$ a calibrated churn probability estimator. For customers with $f(\mathbf{x}) \geq \theta$ (threshold $\theta = 0.5$), CARRE produces the following structured output tuple. Below-threshold customers may receive a reason label for diagnostic evaluation but do not receive a Stage~2 prescription or Stage~3-b explanation:
\begin{equation}
  \text{CARRE}(\mathbf{x}) = (\underbrace{r}_{\text{reason}},\; \underbrace{a^*}_{\text{prescription}},\; \underbrace{e}_{\text{explanation}}),
  \label{eq:carre}
\end{equation}
where $r \in \mathcal{R}$ is a structured churn label, $a^* \in \mathcal{A}$ is a predefined candidate action, and $e$ is a natural-language justification.

The output tuple is designed to satisfy three simultaneous constraints. $r$ must be derivable from $\mathbf{x}$ under the priority-ordered rule taxonomy; $a^*$ must maximize CF risk reduction net of cost among retrieved candidates; and $e$ must be logically consistent with both $r$ and $a^*$. The last constraint is structurally encouraged by Stage~3-b, which conditions generation on both $r$ and $a^*$; consistency is subsequently evaluated rather than guaranteed.

\subsection{Three-Stage Pipeline}

In Stage 1, each action is represented as a document containing an ID, description, target customer profile, expected effect, and eligibility conditions. For example, the \texttt{A\_CONTRACT\_24M} document targets ``month-to-month customers with medium-to-high monthly charges showing interest in long-term commitment.'' The expected-effect text in these documents (e.g., a stated churn reduction of 0.25--0.40) is heuristic metadata used only for retrieval context; it is not treated as a causal or empirically validated effect estimate, and Stage~2 recomputes predicted-risk changes directly from the classifier. Documents are encoded with \texttt{sentence-transformers/all-MiniLM-L6-v2} and stored in a FAISS flat inner-product index \cite{johnson2019billion}. For each customer, a retrieval query is constructed from four key features (\texttt{Contract}, \texttt{PaymentMethod}, \texttt{InternetService}, \texttt{MonthlyCharges}), and the top-$k$ most similar documents (default $k{=}5$) are returned as the candidate action set.

In Stage 2, for each retrieved action $a_i$, its feature modification rules are applied to $\mathbf{x}$ to construct a CF customer vector $\mathbf{x}'_i$; the exact transformations and eligibility conditions are listed in Appendix~\ref{app:actions} (Table~\ref{tab:transforms}). The CF score is:
\begin{equation}
  \text{CF-score}(a_i) = \underbrace{\bigl[f(\mathbf{x}) - f(\mathbf{x}'_i)\bigr]}_{\Delta\text{risk}} - \lambda \cdot \frac{\text{cost}(a_i)}{\max_j \text{cost}(a_j)},
  \label{eq:cf}
\end{equation}
$a^* = \arg\max_{a_i} \text{CF-score}(a_i)$ is selected as the \texttt{cf\_action}. The hyperparameter $\lambda$ controls the risk-cost trade-off and is set to $\lambda{=}0.25$ based on the $\lambda$ sweep in Table~\ref{tab:exp4}. If no retrieved action achieves $\Delta\text{risk} > 0$, Stage~2 returns \texttt{cf\_action=NONE}.

Stage 3 comprises two sequential sub-steps that together connect Stage~1's semantic retrieval and Stage~2's cost-optimal prescription.

\textbf{Stage~3-a (Reason Classification and Auxiliary Plan Generation).} Stage~3-a and Stage~2 operate \emph{in parallel}: both receive the top-$k$ action candidates from Stage~1 independently. The LLM receives a JSON customer feature profile, the baseline churn probability $f(\mathbf{x})$, and the top-$k$ retrieved action documents. Critically, \texttt{cf\_action} is \emph{not} passed to Stage~3-a. Early experiments showed that including the CF-selected action causes the LLM to anchor its explanation backward from the prescribed action rather than classifying the customer profile against the explicit criteria---a pattern we term \emph{reverse-reasoning bias}. Excluding \texttt{cf\_action} enforces the \emph{direction} of inference: from customer features through the Step 1--6 priority rules to a reason label. This provides only \emph{procedural} independence from the prescription---the reason label is not back-derived from \texttt{cf\_action}---and not \emph{epistemic} independence: because the Step 1--6 criteria are explicitly supplied in the prompt, Stage~3-a's task reduces to faithfully applying a fixed decision procedure to the customer profile rather than performing open-ended domain reasoning about churn drivers (see Section~\ref{sec:exp_llm}). This parallel design allows Stage~2 and Stage~3-a to be independently evaluated and improved. Stage~3-a also emits an auxiliary \texttt{llm\_plan} and brief rationale bullets, used only to evaluate constrained instruction following; they are not delivered as the final prescription or explanation. The deployed prescription is the \texttt{cf\_action} from Stage~2, and the delivered explanation is regenerated in Stage~3-b from the Stage~3-a reason and the Stage~2 action.

The Stage~3-a prompt incorporates three design choices. First, explicit Step 1--6 classification criteria are provided, anchoring the LLM to a structured decision procedure. Second, CoT instructions require the LLM to evaluate each step in priority order before committing to a label. Third, a fixed plain-text output schema consisting of a \texttt{Reason:} line, a \texttt{Plan:} line, and two rationale bullets is enforced; plan IDs outside the retrieved set are flagged as auxiliary-plan constraint violations. A condensed prompt template illustrating these choices is provided in Appendix~\ref{app:prompt}.

\textbf{Stage~3-b (Explanation Generation).} Given the reason $r$ from Stage~3-a and the optimal action $a^*$ from Stage~2, Stage~3-b generates the natural-language explanation $e$. The LLM receives the customer profile, $f(\mathbf{x})$, $r$, and $a^*$, and is instructed to provide a profile-grounded rationale---in 2--3 English sentences---for why $a^*$ is appropriate for this customer's identified churn risk $r$, grounding the explanation in specific profile fields. Because Stage~3-b conditions generation on both $r$ and $a^*$, the resulting explanation is intended to align with the actual prescription from Stage~2; alignment is evaluated empirically rather than guaranteed. This two-step design targets the incoherence that arises when Stage~2 and Stage~3-a select different actions, which is the common case at scale (Section~\ref{sec:exp_scale}).

The \texttt{cf\_action} (Stage~2) and \texttt{llm\_plan} (Stage~3-a) may differ, as the two stages optimize for distinct objectives: cost-adjusted risk reduction versus rule-consistent classification. When they diverge, the discrepancy serves as a diagnostic signal---indicating either that cost parameters need adjustment or that rule criteria require refinement---while Stage~3-b conditions the delivered explanation on the actual prescription. As a result, CARRE simultaneously produces \emph{why} a high-risk customer is likely to churn ($r$), \emph{what action} to take ($a^*$), and \emph{why that action} is appropriate ($e$), bridging the prediction-prescription gap that characterizes most deployed churn systems.

%% ─────────────────────────────────────────────────────────────────────
\section{Experimental Setup}
\label{sec:experiments}
%% ─────────────────────────────────────────────────────────────────────

We organize the evaluation around three research questions:
\begin{description}[leftmargin=0pt,labelindent=0pt,font=\normalfont\bfseries]
  \item[RQ1.] Does direct counterfactual simulation produce more effective and cost-efficient prescriptions than random, rule-based, and SHAP-based alternatives?
  \item[RQ2.] How do retrieval depth $k$ and the cost penalty $\lambda$ affect candidate coverage, prescription quality, and operational cost?
  \item[RQ3.] How reliably do LLMs apply the reason taxonomy and generate explanations that are coherent, constrained, and grounded in customer evidence?
\end{description}

We use the IBM Telco Customer Churn dataset \cite{ibm2019telco}. After removing the customer identifier and the churn label, each of the 7,043 customers is described by 19 predictor features covering demographics (SeniorCitizen, gender), service subscriptions (InternetService, TechSupport, OnlineSecurity), contract conditions (Contract, tenure, MonthlyCharges), and payment method. The binary churn label has a positive class prevalence of 26.5\%, representing moderate class imbalance. We apply an 80/20 stratified split with seed 42, yielding 5,634 training and 1,409 test instances.

All experiments are conducted in a Python environment using scikit-learn, FAISS, sentence-transformers, and the OpenAI/Groq API; preprocessing, hyperparameters, model identifiers, and seed settings are summarized in Appendix~\ref{app:repro}. The CF prescription experiments (Tables~\ref{tab:exp2}--\ref{tab:exp4}) use all 313 high-risk test cases ($f(\mathbf{x}) \geq 0.50$) in the test split, while the LLM reasoning experiments use a 30-case balanced set (Table~\ref{tab:exp5}) for model comparison and a scaled 136-case reason-stratified sample (Tables~\ref{tab:judge},~\ref{tab:exp7}) for the adopted model.

CARRE's Stage 2 CF scoring requires a reliable churn probability estimate $f(\mathbf{x})$. We use a \texttt{CalibratedClassifierCV}-wrapped \texttt{LogisticRegression} (C=1.0, max\_iter=2000) with balanced class weighting and sigmoid calibration under 3-fold cross-validation. Although tree-based ensembles such as gradient boosting or random forests can match or slightly exceed logistic regression on aggregate ranking metrics, the isotonic-calibrated tree ensembles tested in our implementation exhibited coarse and discontinuous probability responses under graded feature perturbations, making it difficult to estimate the subtle risk changes on which CF simulation depends. Logistic regression instead provides a smooth, continuous probability surface that responds gradually to feature value changes, making it better suited for CF analysis where we directly estimate how much a given action reduces churn probability. We validate this trade-off empirically in Appendix~\ref{app:model} (Table~\ref{tab:model_cf}). Table~\ref{tab:exp1} reports three performance metrics on the test set: ROC-AUC measures discriminative ability under class imbalance, PR-AUC captures precision-recall balance on the minority class, and Brier Score quantifies calibration quality.

\begin{table}[t]
\centering
\caption{Base prediction model performance (IBM Telco Churn, 1,409 test instances).}
\label{tab:exp1}
\begin{tabular}{lc}
\toprule
Metric & Value \\
\midrule
ROC-AUC     & 0.842 \\
PR-AUC      & 0.633 \\
Brier Score & 0.138 \\
\bottomrule
\end{tabular}
\end{table}

ROC-AUC of 0.842 confirms sufficient discriminative performance for imbalanced binary classification. PR-AUC of 0.633 is substantially above the random baseline (0.265) given a positive class prevalence of 26.5\%. A Brier score of 0.138 indicates reasonable probabilistic accuracy on this test split, although a full calibration assessment would also require reliability curves or calibration-error measures. This probabilistic accuracy directly affects the Stage~2 CF scores.

\textbf{Base classifier choice for CF scoring.} A natural question is why CARRE retains logistic regression rather than a tree ensemble calibrated via isotonic regression, which attains marginally higher ROC-AUC (0.845 for gradient boosting vs.\ 0.842). A graded-intervention sweep on the 313 high-risk cases shows that all three tested tree ensembles produce severely non-smooth CF surfaces: 45--66\% of the intervention range is completely flat and single-step jumps reach 0.06--0.10, so small realistic interventions register either as zero risk change or as an over-large discontinuous drop. Logistic regression instead yields a fully continuous, gradually responding surface (0\% flat, maximum jump 0.002), producing the graded $\Delta$risk estimates that Stage~2 requires. Sacrificing this CF reliability for a 0.003 ROC-AUC gain is an unfavorable trade, so we retain calibrated logistic regression. The comparison protocol and full per-model results are given in Appendix~\ref{app:model} (Table~\ref{tab:model_cf}).

\section{Results}
\label{sec:results}

\subsection{Prescription Quality and Cost Efficiency}
\label{sec:exp_pipeline}

This section evaluates CARRE's prescription pipeline in three aspects in sequence: prescription quality against baselines, RAG retrieval depth ($k$), and the risk-cost balance hyperparameter ($\lambda$). All experiments are conducted on all 313 high-risk test cases ($f(\mathbf{x}) \geq 0.50$), the complete high-risk cohort of the test split rather than a truncated sample. For the prescription-method comparison in Table~\ref{tab:exp2}, per-case risk-reduction metrics are reported with 95\% bootstrap confidence intervals.

To compare prescription methods, CARRE is evaluated against four baselines. The random baseline uniformly samples one eligible action for each customer under a fixed per-customer seed. The rule-based baseline applies a priority-ordered profile-to-action lookup table. The SHAP-based baseline maps the highest positive churn-directed SHAP feature to an eligible candidate action. The \emph{SHAP+Cost} baseline selects among SHAP-implicated eligible actions using the same cost-penalty structure as CARRE's Equation~\ref{eq:cf}, with normalized SHAP importance replacing the model-predicted risk-reduction term. The exact rule-based and SHAP-based mappings are given in Appendix~\ref{app:baselines} (Table~\ref{tab:baselines}). Table~\ref{tab:exp2} compares five methods across four metrics: Risk Red.\ is the mean model-predicted churn-probability reduction per case, Effectiveness is the fraction of cases achieving $\Delta\text{risk} > 0$, Avg.\ Rel.\ Cost is the mean researcher-defined relative action cost, and Efficiency is the ratio of Risk Red.\ to Avg.\ Rel.\ Cost.

\begin{table}[t]
\centering
\caption{Prescription method comparison (all 313 high-risk cases with $f(\mathbf{x}){\geq}0.5$, $\lambda{=}0.25$). Brackets denote 95\% bootstrap confidence intervals (2{,}000 resamples). SHAP+Cost applies the identical cost penalty used in CARRE's CF-score to SHAP-selected actions.}
\label{tab:exp2}
\small
\begin{adjustbox}{max width=\linewidth}
\begin{tabular}{lccccc}
\toprule
Method & Risk Red.$\uparrow$ & 95\% CI & Effectiveness$\uparrow$ & Avg. Rel. Cost & Efficiency$\uparrow$ \\
\midrule
Random      & 0.137 & [0.125, 0.150] & 92.3\%  & 2.26 & 0.061 \\
Rule-based  & 0.074 & [0.064, 0.085] & 48.9\%  & 2.51 & 0.030 \\
SHAP-based  & 0.168 & [0.166, 0.169] & 99.7\% & 2.00 & 0.084 \\
SHAP+Cost   & 0.167 & [0.165, 0.169] & 99.4\% & 2.01 & 0.083 \\
\textbf{CARRE} & \textbf{0.302} & [0.294, 0.308] & \textbf{100\%} & 3.26 & \textbf{0.093} \\
\bottomrule
\end{tabular}
\end{adjustbox}
\end{table}

CARRE outperforms all baselines on risk reduction and efficiency. Against the strongest baseline, CARRE achieves 79.8\% higher raw risk reduction than the SHAP-based approach (0.302 vs.\ 0.168), with non-overlapping 95\% confidence intervals ([0.294, 0.308] vs.\ [0.166, 0.169]). Because CARRE also selects higher-cost actions on average (Avg.\ Rel.\ Cost 3.26 vs.\ 2.00), we report two complementary controlled comparisons. First, a cost-normalized comparison using Efficiency yields a more conservative 10.5\% advantage (0.093 vs.\ 0.084). Second, and more directly, the SHAP+Cost baseline subjects SHAP-based selection to the identical cost penalty used by CARRE. SHAP+Cost yields nearly the same mean predicted-risk reduction as plain SHAP (0.167 vs.\ 0.168), and their marginal bootstrap intervals overlap; CARRE exceeds SHAP+Cost by 80.4\% in mean predicted-risk reduction. Together these comparisons show that CARRE's advantage stems from direct CF simulation rather than from cost weighting---penalizing cost does not help feature-importance-based selection, because SHAP importance for prediction does not identify the actions that most reduce churn risk.

Notably, the high effectiveness of the random baseline (92.3\%) reflects a design property of the action space rather than a CARRE-specific achievement: all six actions were deliberately chosen as business interventions targeting high-risk churn customers, making most of them broadly applicable across this population regardless of selection method. Effectiveness ($\Delta\text{risk} > 0$) is therefore a binary coverage metric capturing whether any positive effect is achieved, not the magnitude of that effect. The more discriminating measures of method quality are Risk Red.\ and Efficiency, where CARRE substantially outperforms all baselines: CARRE achieves 120\% higher risk reduction than random (0.302 vs.\ 0.137) and 52\% higher efficiency (0.093 vs.\ 0.061), confirming that what differentiates methods is not whether they find an effective action, but how much risk reduction they deliver per unit cost.

CARRE's advantage on these quality metrics stems from direct CF simulation. The SHAP-based approach targets features with high explanation importance, but importance for prediction does not necessarily coincide with the highest CF risk reduction. CARRE instead directly simulates each action as a CF intervention, measures the resulting churn probability reduction, and selects the optimal action after balancing cost. The rule-based baseline scores lowest on both risk reduction (0.074) and effectiveness (48.9\%) because the static lookup table handles only one rule at a time, failing to identify the correct action for customers with multiple simultaneous risk factors; its low effectiveness is attributable to rule mismatch, not action space limitations.

Turning to the effect of RAG retrieval depth $k$, Table~\ref{tab:exp3} reports four metrics: Recall@$k$ is the fraction of cases where the globally optimal action is within the top-$k$ retrieved set, Risk Red.\ is the mean CF risk reduction, Avg.\ Rel.\ Cost is the mean researcher-defined relative action cost, and CF-NONE Rate is the fraction of cases for which Stage~2 returns \texttt{cf\_action=NONE}.

\begin{table}[t]
\centering
\caption{Effect of retrieval depth $k$ on CF performance (313 high-risk cases).}
\label{tab:exp3}
\small
\begin{adjustbox}{max width=\linewidth}
\begin{tabular}{lcccc}
\toprule
$k$ & Recall@$k$$\uparrow$ & Risk Red.$\uparrow$ & Avg. Rel. Cost & CF-NONE Rate$\downarrow$ \\
\midrule
1 & 0.000 & 0.028 & 1.275 & 48\% \\
2 & 0.006 & 0.115 & 1.764 &  7\% \\
3 & 0.013 & 0.165 & 1.949 &  0\% \\
4 & 0.709 & 0.277 & 3.022 &  0\% \\
\textbf{5} & \textbf{0.853} & \textbf{0.302} & \textbf{3.260} & \textbf{0\%} \\
6 (oracle) & 1.000 & 0.327 & 3.500 &  0\% \\
\bottomrule
\end{tabular}
\end{adjustbox}
\end{table}

The results reveal a clear retrieval threshold effect. The sharp recall jump from 1.3\% to 70.9\% as $k$ increases from 3 to 4 occurs because the retrieval query represents the customer's current state, so status-quo-similar documents tend to rank first through third, while high-impact actions (e.g., \texttt{A\_CONTRACT\_24M} for month-to-month customers) tend to appear at rank 4 or lower. At $k{=}5$, recall reaches 85.3\% and NONE cases are eliminated; the 14.7\% recall gap versus the oracle ($k{=}6$) is offset by the benefit of reducing context noise passed to the LLM. We adopt $k{=}5$ as the default.

Table~\ref{tab:exp4} summarizes the $\lambda$ sweep, where $\lambda$ (Equation~\ref{eq:cf}) governs the risk--cost balance. Risk Red.\ (mean $\Delta\text{risk}$ per case) and Avg.\ Rel.\ Cost (mean relative intervention cost) reflect the two objectives of Equation~\ref{eq:cf} \cite{karimi2021algorithmic,mothilal2020explaining}; Efficiency is their ratio (a study-specific composite, not a standard CF metric); and High-Cost Rate is the fraction of cases selecting an action with relative cost $\geq 3.0$, included to gauge practical feasibility.

\begin{table}[t]
\centering
\caption{$\lambda$ sweep: risk-cost trade-off (313 high-risk cases, $k{=}5$).}
\label{tab:exp4}
\small
\begin{adjustbox}{max width=\linewidth}
\begin{tabular}{lcccc}
\toprule
$\lambda$ & Risk Red.$\uparrow$ & Avg. Rel. Cost & Efficiency$\uparrow$ & High-Cost Rate \\
\midrule
0.10 & 0.303 & 3.280 & 0.0923 & 85\% \\
\textbf{0.25} & \textbf{0.302} & \textbf{3.260} & \textbf{0.0925} & 85\% \\
0.50 & 0.122 & 1.393 & 0.0878 & 0\% \\
1.00 & 0.103 & 1.272 & 0.0812 & 0\% \\
1.50 & 0.103 & 1.272 & 0.0812 & 0\% \\
\bottomrule
\end{tabular}
\end{adjustbox}
\end{table}

The sweep reveals a sharp regime boundary rather than a gradual trade-off. For $\lambda{\in}[0.10, 0.25]$, the cost penalty is small relative to the risk-reduction gap between actions, so the highest-impact action (\texttt{A\_CONTRACT\_24M}) dominates selection and risk reduction is essentially flat (0.303 vs.\ 0.302, a 0.4\% difference). At $\lambda{=}0.50$ the penalty overtakes that gap, selection flips almost entirely to low-cost actions (\texttt{A\_PAYMENT\_AUTOPAY}), and risk reduction collapses by 59\% (0.302 $\to$ 0.122). Beyond $\lambda{=}1.0$ the sweep saturates. $\lambda{=}0.25$ therefore sits at the knee of this transition: it attains near-maximum risk reduction (within 0.4\% of the $\lambda{=}0.10$ peak) with Efficiency essentially tied with $\lambda{=}0.10$ (0.0925 vs.\ 0.0923), while any further increase crosses the regime boundary and forfeits over half the achievable risk reduction. We adopt $\lambda{=}0.25$ as the default. The discreteness of this transition follows from the small six-action space with widely separated costs, and its dataset-specific boundary is discussed as a limitation in Section~\ref{sec:discussion}. In practice, $\lambda$ is domain-calibratable: organizations with strict cost ceilings should operate at $\lambda \geq 0.5$, others below the boundary.

\textbf{Retrieval-only action-document scaling stress test.} A natural deployment question is how Stage~1 retrieval behaves as the action catalog grows from six curated actions to hundreds of dynamic marketing levers. We ran a retrieval-only stress test of Stage~1---the downstream CF and LLM stages were not executed---augmenting the six action documents with up to 394 synthetic lever documents (retrieval documents, not executable actions) and measuring retrieval runtime and oracle-action survival over all 313 high-risk cases. Runtime is not a bottleneck: FAISS search itself stays below 0.01\,ms per query even at $|\mathcal{A}|{=}400$, and end-to-end query latency, dominated by embedding, stays near 1\,ms. Retrieval precision, however, degrades in two stages---exact-action Recall@5 collapses once near-duplicate parameter variants appear, and family-level recall collapses beyond $|\mathcal{A}|{\approx}200$. Scaling to industrial catalogs therefore requires hierarchical or diversity-aware retrieval rather than a larger flat $k$. The full protocol and results are given in Appendix~\ref{app:scaling} (Table~\ref{tab:scaling}).

\subsection{LLM Reason Classification}
\label{sec:exp_llm}

This subsection evaluates how reliably LLMs apply the reason taxonomy across models; the two that follow assess explanation quality at scale and the effect of retrieval scope. \textbf{Agreement} refers to the rate at which LLM output labels match the weak labels determined by the Step 1--6 priority rules in Section~\ref{sec:framework}. Because the same rules are supplied in the prompt, this metric measures \emph{instruction-following fidelity}---how consistently the LLM applies a given decision procedure---rather than independent domain reasoning ability.

The LLM model comparison was conducted using an earlier model-comparison prompt with the CF action withheld, explicit classification criteria, and CoT instructions. This formulation differs from the v2 scaled-evaluation prompt used for the 136-case experiments in Sections~\ref{sec:exp_scale} and~\ref{sec:exp_rag}. Table~\ref{tab:exp5} compares four LLMs across three metrics: Agreement is the weak-label agreement rate, F1-macro is the macro-averaged F1 across six reason labels, and auxiliary-plan violations are the fraction of cases whose Stage~3-a \texttt{llm\_plan} falls outside the retrieved set. A per-reason breakdown is provided in Appendix~\ref{app:reason} (Table~\ref{tab:exp5_reason}).

\begin{table}[t]
\centering
\caption{LLM comparison on reason classification (model-comparison prompt; 30-case balanced set, 5 per reason category). Because this formulation differs from the v2 scaled-evaluation prompt, its 0.933 agreement is not directly comparable to the 136-case results in Sections~\ref{sec:exp_scale} and~\ref{sec:exp_rag}.}
\label{tab:exp5}
\small
\begin{adjustbox}{max width=\linewidth}
\begin{tabular}{llccc}
\toprule
Model & Type & Agreement$\uparrow$ & F1-macro$\uparrow$ & Aux.-plan viol.$\downarrow$ \\
\midrule
Qwen3-32B (Groq)     & OSS (L)   & 0.367 & 0.321 & 0.000 \\
GPT-4o-mini          & Prop.\ (S) & 0.400 & 0.301 & 0.000 \\
Llama-3.3-70B (Groq) & OSS (L)   & 0.500 & 0.512 & 0.000 \\
\textbf{GPT-4o}      & Prop.\ (L) & \textbf{0.933} & \textbf{0.933} & \textbf{0.000} \\
\bottomrule
\end{tabular}
\end{adjustbox}
\end{table}

GPT-4o achieves the highest agreement (93.3\%), with two errors out of 30. One falls in CONTRACT\_RISK (4/5), where a customer with tenure$\leq$3 satisfies both Steps 2 and 5 and GPT-4o inverts the priority order; the other falls in OTHER (4/5). GPT-4o-mini (40.0\%) shows a distinct avoidance pattern, predicting 0\% on PRICE\_SENSITIVE and CONTRACT\_RISK while achieving 100\% on PAYMENT\_FRICTION and 80\% on OTHER, suggesting it anchors heavily on surface payment-method cues while failing on price- and contract-related rules. Llama-3.3-70B (50.0\%) achieves 100\% on PRICE\_SENSITIVE but scores zero on CONTRACT\_RISK and OTHER. Llama-3.3-70B outperforms GPT-4o-mini in this small comparison, but the models differ simultaneously in scale, architecture, vendor, and training, so the result does not isolate which factor drives the difference. Qwen3-32B (36.7\%) shows the most severe bias, scoring zero on CONTRACT\_RISK, PAYMENT\_FRICTION, and OTHER; for CONTRACT\_RISK this reflects systematic \emph{misclassification} rather than abstention, as the model over-predicts that label yet never applies it correctly, whereas for PAYMENT\_FRICTION and OTHER it abstains from the label entirely. This pattern is consistent with the known sensitivity of prompted language models to label and contextual biases~\cite{zhao2021calibrate}, although the current experiment does not identify its cause. Qwen3-32B was evaluated with its built-in Extended Thinking mode disabled to apply the same external CoT prompt uniformly across all models. No auxiliary-plan constraint violations occurred for any of the four models in this 30-case comparison. We adopt GPT-4o as the default LLM for subsequent experiments. It is then evaluated on a larger 136-case sample using the v2 scaled-evaluation prompt. Because the prompt formulation also changes, this evaluation should not be interpreted as a pure sample-size comparison with Table~\ref{tab:exp5}.

\subsection{Explanation Quality at Scale}
\label{sec:exp_scale}

For human evaluation of generated explanations, four independent raters assessed GPT-4o outputs on a 1--5 Likert scale across four criteria: Faithfulness (alignment with the customer's actual data), Actionability (perceived operational feasibility of the recommended action), Coherence (logical consistency between the stated reason and recommended plan), and Satisfaction (overall perceived quality). However, inter-rater agreement was very low (Krippendorff's $\alpha=0.003$--$0.175$), indicating that the ratings were insufficiently reliable for confirmatory analysis. We therefore do not treat the per-criterion human ratings as findings; a rigorous redesign with structured rater calibration, explicit criterion operationalization, and $n \geq 25$ cases per reason category would be required before human evaluation can serve as confirmatory evidence \cite{artstein2008inter}.

In place of the discarded human study, we conducted a scaled automated evaluation on a reason-stratified sample of 136 cases drawn from the test split (25 per reason category; 11 for \texttt{LOW\_ENGAGEMENT})---a 4.5$\times$ expansion over the 30-case set. Each case was processed by the full pipeline with the v2 reason prompt, and each explanation was scored 1--5 on the three criteria by two cross-vendor LLM judges (GPT-4o, Llama-3.3-70B) under a G-Eval-style protocol \cite{liu2023geval} with explicit criterion definitions and mandated field-by-field verification (Table~\ref{tab:judge}). The explanation evaluation therefore reflects the pre-refinement v2 reason outputs. Three findings emerge. First, both judges rate explanation quality high (faithfulness 4.02/4.04, coherence 4.24/4.47, actionability 4.45/5.00), with within-one-point cross-vendor agreement of 61--87\%. The Llama judge saturates on actionability (5.00), so its scores on that criterion carry no discriminative signal. Second, a deterministic grounding check---verifying every categorical and numeric claim against the actual profile with no LLM involvement---found zero contradictions across the 66 verifiable claims present in 55 of the 135 explanations; the other 80 state no checkable claim, and both judges returned valid scores for all 135. A claim is verifiable when it asserts a contract type, payment method, dollar charge, or tenure checkable against the profile. Third, under the same v2 scaled-evaluation prompt, agreement decreases modestly from 83.3\% on the 30-case ablation set (Table~\ref{tab:exp7}) to 79.4\% on the 136-case sample. On the larger sample, errors concentrate in \texttt{CONTRACT\_RISK} and \texttt{OTHER}: \texttt{CONTRACT\_RISK} reaches 52\% because of Step~2 versus Step~5/Step~3 priority inversions, while \texttt{OTHER} reaches 48\% because of over-application of the \texttt{SUPPORT\_DEFICIT} rule. The 93.3\% result in Table~\ref{tab:exp5} used a different model-comparison prompt and is therefore not directly comparable. Because both failure patterns were addressable through prompt changes on this sample, we added a targeted enforcement block---lowest-step-wins tie-breaking with worked examples, and an explicit all-three-conditions requirement for Step~4. This \emph{v3} prompt increased agreement from 79.4\% to 90.4\% (F1-macro 0.910, zero violations) on the same 136 cases: \texttt{CONTRACT\_RISK} recovers to 100\% and \texttt{OTHER} to 88\%, at the cost of partial over-correction on \texttt{SUPPORT\_DEFICIT} (60\%). This post-refinement result may be optimistic because the sample was used for both error diagnosis and re-evaluation, and it therefore requires confirmation on an independent held-out set. We adopt v3 as CARRE's final configuration and retain 79.4\% as the pre-refinement diagnostic estimate. Zero grounding contradictions coexist with 79.4\% label agreement because the grounding check verifies factual claims against the profile, not the upstream reason label (Case~3, Section~\ref{sec:case}). These LLM judges share training-distribution biases with the generator and do not replace calibrated human evaluation; they serve as scalable interim evidence pending a redesigned human protocol with rater calibration and per-category coverage.

\begin{table}[t]
\centering
\caption{Automated explanation evaluation on a 136-case reason-stratified subset of the test split (135 generated explanations); explanations use the v2 reason outputs. Judges score 1--5 under a G-Eval-style protocol; grounding is a deterministic claim-vs-profile check involving no LLM.}
\label{tab:judge}
\small
\begin{adjustbox}{max width=\linewidth}
\begin{tabular}{lccc}
\toprule
Criterion & GPT-4o judge & Llama-3.3-70B judge & Within-1 agr. \\
\midrule
Faithfulness  & 4.02 & 4.04 & 61\% \\
Coherence     & 4.24 & 4.47 & 78\% \\
Actionability & 4.45 & 5.00 & 87\% \\
\midrule
Grounding accuracy & \multicolumn{3}{c}{66/66 verifiable claims consistent (100\%)} \\
\bottomrule
\end{tabular}
\end{adjustbox}
\end{table}

\paragraph{Stage 3-b: Reason--Action Concordance and Explanation Quality.}
Stage~3-b was applied to the 136-case reason-stratified sample to generate explanations aligned with the Stage~2 \texttt{cf\_action}. The Stage~3-a auxiliary \texttt{llm\_plan} and the Stage~2 \texttt{cf\_action} match exactly in 2\% of all 135 explained cases (3/135). Among the 89 cases where both stages select an action, exact match is 3\% (3/89) and action-family match is 11\% (10/89), because \texttt{A\_CONTRACT\_24M} dominates Stage~2 selections (99/135) wherever switching to a two-year contract gives the highest cost-penalized CF score. Part of the divergence is structural: 47 of the 136 cases have Stage~3-a \texttt{plan=NONE}. Stage~2 nevertheless finds a positive-$\Delta$risk \texttt{cf\_action} for 46 of these cases; the remaining case has no CF action, yielding 135 explanations. Stage~3-b generates direct profile-grounded explanations for concordant pairs and bridging rationales for divergent ones (Case~2, Section~\ref{sec:case}). This divergence does not render Stage~3-a redundant: the reason label is not a prescription input but serves three roles---grounding the Stage~3-b explanation, framing the agent-facing conversation, and flagging cost-dominance for $\lambda$ or action-space recalibration.

\subsection{Effect of Retrieval Scope}
\label{sec:exp_rag}

To isolate RAG's contribution to LLM reasoning beyond its established role of constraining the plan space, GPT-4o was run under three conditions. Table~\ref{tab:exp7} compares three conditions on Agreement (weak-label agreement rate), F1-macro (macro-averaged F1), and auxiliary-plan violations (fraction of Stage~3-a \texttt{llm\_plan} selections outside the retrieved set; these do not affect the final \texttt{cf\_action}).

\begin{table}[t]
\centering
\caption{Effect of retrieval scope on LLM reasoning quality (GPT-4o), at the original 30-case set and re-run on the 136-case sample. The $k{=}3$ peak observed at $n{=}30$ does not replicate at scale; the $k{=}5$ default is best at $n{=}136$. F1-macro and auxiliary-plan violations are reported for the $n{=}136$ run only. The 136-case retrieval ablation uses the v2 reason prompt, which is distinct from the model-comparison formulation in Table~\ref{tab:exp5}, so its $k{=}5$/30-case agreement (0.833) is not directly comparable to that table's 0.933.}
\label{tab:exp7}
\small
\begin{adjustbox}{max width=\linewidth}
\begin{tabular}{lcccc}
\toprule
 & \multicolumn{2}{c}{Agreement$\uparrow$} & F1-macro$\uparrow$ & Aux.-plan viol.$\downarrow$ \\
\cmidrule(lr){2-3}
Condition & $n{=}30$ & $n{=}136$ & ($n{=}136$) & \\
\midrule
No-RAG ($k{=}6$, all actions) & 0.767 & 0.757 & 0.746 & 0.000 \\
RAG $k{=}3$                   & 0.967 & 0.765 & 0.738 & 0.000 \\
RAG $k{=}5$ (default)         & 0.833 & \textbf{0.794} & \textbf{0.782} & 0.000 \\
\bottomrule
\end{tabular}
\end{adjustbox}
\end{table}

On the original 30-case set, $k{=}3$ appeared to peak dramatically (96.7\%), suggesting that tighter retrieval might outperform the $k{=}5$ default. Re-running the ablation on the 136-case sample shows that this peak was a small-sample artifact: at scale, $k{=}3$ (76.5\%) is essentially level with No-RAG (75.7\%), and the CARRE default $k{=}5$ is best on both agreement (79.4\%) and F1-macro (0.782). Two conclusions replace the earlier reading. First, constraining the LLM's context to retrieved actions still helps---No-RAG trails $k{=}5$ by 3.7 percentage points, consistent with irrelevant actions inducing \emph{action-to-reason reverse inference}---but the effect is modest rather than dramatic. Second, $k{=}5$ provides the best evaluated compromise: it attains high, though not maximal, oracle-action coverage---the full-library $k{=}6$ condition reaches maximal coverage (Table~\ref{tab:exp3})---together with the highest downstream classification agreement, at zero violations in every condition.

%% ─────────────────────────────────────────────────────────────────────
\subsection{Qualitative Case Study}
\label{sec:case}
%% ─────────────────────────────────────────────────────────────────────

Table~\ref{tab:case} presents three high-risk outputs from the CARRE prescription-evaluation cohort and one low-risk abstention control (GPT-4o, $k{=}5$). The low-risk control is included only to illustrate threshold-based non-intervention and is not part of the 313-case high-risk prescription evaluation.

\begin{table}[t]
\centering
\caption{Case study: concordant success (top), Stage~3-b mismatch bridge (second), step priority failure (third), and low-risk abstention control (bottom). Per-case scores are from the two-judge automated evaluation of Section~\ref{sec:exp_scale}, shown as G:$x$/L:$y$ for the GPT-4o and Llama-3.3-70B judges (1--5 scale); scores from the discarded human study are not reported. Concordance is defined at the level of action family (e.g., contract-switch), not exact action ID.}
\label{tab:case}
\footnotesize
\begin{tabular}{p{1.9cm}p{5.6cm}}
\toprule
\multicolumn{2}{l}{\textbf{Case 1: Concordant Success (Customer 2514-GINMM)}} \\
\midrule
Churn Prob.       & 0.741 \\
Profile           & Contract: M-t-M, tenure: 4 mo., charges: \$65.7 \\
CF Action (S2)    & \texttt{A\_CONTRACT\_24M} ($\Delta$risk: 0.326) \\
3-a Reason / Aux. Plan & \texttt{CONTRACT\_RISK} / \texttt{A\_CONTRACT\_12M} \\
3-b Explanation   & ``Upgrading to a 2-Year Contract addresses the contract instability from the month-to-month arrangement and short tenure.'' \\
LLM Judges        & Faith.\ G:5/L:5 $\cdot$ Coh.\ G:5/L:5 $\cdot$ Action.\ G:5/L:5 \\
\midrule
\multicolumn{2}{l}{\textbf{Case 2: Stage~3-b Mismatch Bridge (Customer 2077-MPJQO)}} \\
\midrule
Churn Prob.       & 0.741 \\
Profile           & Contract: M-t-M, tenure: 7 mo., charges: \$75.4, pay.: Electronic check \\
CF Action (S2)    & \texttt{A\_CONTRACT\_24M} ($\Delta$risk: 0.330) \\
3-a Reason / Aux. Plan & \texttt{PAYMENT\_FRICTION} / \texttt{A\_PAYMENT\_AUTOPAY} \\
3-b Explanation   & ``A 2-Year Contract addresses payment friction with a more stable, predictable billing cycle.'' \\
LLM Judges        & Faith.\ G:3/L:3 $\cdot$ Coh.\ G:3/L:3 $\cdot$ Action.\ G:4/L:5 \\
\midrule
\multicolumn{2}{l}{\textbf{Case 3: Step Priority Error (Customer 8258-GSTJK)}} \\
\midrule
Churn Prob.       & 0.756 \\
Profile           & Contract: M-t-M, tenure: 3 mo., charges: \$69.7 \\
CF Action (S2)    & \texttt{A\_CONTRACT\_12M} ($\Delta$risk: 0.152) \\
3-a Reason / Aux. Plan & \texttt{LOW\_ENGAGEMENT} (correct: \texttt{CONTRACT\_RISK}) / \texttt{NONE} \\
3-b Explanation   & ``Upgrading to a 1-Year Contract suits the customer's low engagement and short tenure, increasing commitment and lowering churn risk.'' \\
LLM Judges        & Faith.\ G:5/L:5 $\cdot$ Coh.\ G:5/L:5 $\cdot$ Action.\ G:5/L:5 \\
\midrule
\multicolumn{2}{l}{\textbf{Case 4: Low-Risk Abstention Control (Customer 1166-PQLGG)}} \\
\midrule
Churn Prob.       & 0.002 \\
Profile           & Contract: 2-yr, tenure: 72 mo., charges: \$19.6, InternetService: No, pay.: bank transfer \\
CF Action (S2)    & NONE ($f(\mathbf{x}){<}0.5$; below intervention threshold) \\
3-a Reason / Aux. Plan & \texttt{OTHER} (correct) / \texttt{NONE} \\
3-b Explanation   & N/A (no CF action to explain) \\
LLM Judges        & N/A (low-risk abstention; no explanation generated) \\
\bottomrule
\end{tabular}
\end{table}

\textbf{Case 1 (concordant success).} Customer 2514-GINMM is correctly classified \texttt{CONTRACT\_RISK}; Stage~2 selects \texttt{A\_CONTRACT\_24M} ($\Delta\text{risk}{=}0.326$) and Stage~3-b coherently links the contract-instability reason to the 24-month prescription. Both judges rate it at ceiling (5/5).

\textbf{Case 2 (mismatch bridge).} Customer 2077-MPJQO is classified \texttt{PAYMENT\_FRICTION}, but Stage~2 selects \texttt{A\_CONTRACT\_24M} ($\Delta\text{risk}{=}0.330$), a far larger model-predicted risk reduction than a payment-method change. Stage~3-b bridges the gap via the contract change, but both judges score it below Case~1 (faithfulness/coherence 3/3), showing that divergence yields less convincing rationale (cf.\ F5).

\textbf{Case 3 (silent priority error).} Customer 8258-GSTJK satisfies both Step~2 and Step~5; the correct label is \texttt{CONTRACT\_RISK} (Step~2 precedence), but GPT-4o applies Step~5 first and misclassifies as \texttt{LOW\_ENGAGEMENT}. Stage~3-b still produces a fluent explanation around the wrong reason, rated 5/5 by both judges---the error surfaces only through the weak-label check, which is why we treat judge scores as surface quality, not end-to-end correctness.

\textbf{Case 4 (low-risk abstention control).} Customer 1166-PQLGG is a long-tenured (72-month), low-charge, phone-only customer on a two-year contract, with a churn probability of only 0.002 and no risk pattern matching Steps 1--5, so the diagnostic reason label is \texttt{OTHER}. Because $f(\mathbf{x})$ falls below the intervention threshold, Stage~2 returns \texttt{cf\_action=NONE} and Stage~3-b generates no explanation. It lies outside the 313-case evaluation cohort and illustrates threshold-based abstention only.

%% ─────────────────────────────────────────────────────────────────────
\section{Discussion and Limitations}
\label{sec:discussion}
%% ─────────────────────────────────────────────────────────────────────

We draw five main findings from the experimental results, alongside four limitations. Finding F1 addresses RQ1, F2 addresses RQ2, and F3--F5 address RQ3.

\textbf{F1: CF optimization outperforms explanation-based heuristics, and the advantage is not an artifact of cost weighting.} CARRE achieves 79.8\% higher raw risk reduction than the SHAP-based approach on all 313 high-risk cases (Table~\ref{tab:exp2}). The controlled \emph{SHAP+Cost} baseline (Section~\ref{sec:exp_pipeline}) rules out the built-in cost penalty as the source of this gap: subjecting SHAP-based selection to the same penalty leaves its risk reduction unchanged, and CARRE retains an 80.4\% advantage. The cost-normalized Efficiency comparison likewise favors CARRE (10.5\%). The gap therefore stems from direct CF simulation, not cost weighting, and shows that prediction feature importance and score-selected intervention targets can diverge---questioning the practice of driving retention campaigns directly from SHAP.

\textbf{F2: Retrieval scope has a modest but real effect on reasoning quality, and small-sample ablations overstate it.} Beyond the plan-space constraint that is RAG's established benefit, restricting the LLM's context to retrieved actions improves weak-label agreement by 3.7 percentage points over the full action library, with zero auxiliary-plan violations throughout. A dramatic $k{=}3$ peak on the 30-case set did not replicate at 136 cases. The $k{=}5$ setting is the best evaluated compromise: high though not maximal oracle-action coverage and the highest weak-label agreement, whereas full-library $k{=}6$ keeps maximal CF coverage but lower reasoning agreement.

\textbf{F3: Prompt engineering is critical for structured LLM classification, and its gains require independent validation.} Withholding the CF-selected action and adding explicit Step~1--6 criteria with CoT improved rule-following during prompt development. Under the same v2 scaled-evaluation prompt, agreement decreased modestly from 83.3\% on the 30-case ablation set to 79.4\% on the 136-case sample. A targeted v3 enforcement block then increased agreement to 90.4\% on the same 136 cases. Because the same cases were used for error diagnosis and re-evaluation, the v3 improvement remains a prompt-development result requiring confirmation on an independent held-out sample. The 93.3\% model-comparison result in Table~\ref{tab:exp5} used a different prompt formulation and is not directly comparable.

\textbf{F4: Structured-reasoning performance varies substantially across models, but the responsible model factors remain confounded.} Llama-3.3-70B reaches 50.0\% and outperforms GPT-4o-mini at 40.0\%, but the pair differs in scale, architecture, and vendor at once, so no single factor is isolated. All three non-GPT-4o models fail systematically in tail categories---Qwen3-32B, for instance, scores zero on \texttt{CONTRACT\_RISK}, \texttt{PAYMENT\_FRICTION}, and \texttt{OTHER}---so the tested open-source models may need targeted calibration before use in this task, pending broader comparison.

\textbf{F5: Cost-aware optimization systematically overrides intuitive rule-based prescriptions, and Stage~3-b bridges the resulting gap.} Across all 135 explained cases, the Stage~3-a auxiliary plan and the Stage~2 action match exactly in 2\% (3/135); among the 89 cases where both stages select an action, exact match is 3\% (3/89) and family-level match is 11\% (10/89), because \texttt{A\_CONTRACT\_24M} dominates Stage~2 selections wherever switching to a two-year contract yields the highest cost-penalized CF score, even for \texttt{PAYMENT\_FRICTION} and \texttt{SUPPORT\_DEFICIT} profiles. The parallel architecture makes this cost-dominance transparent, and Stage~3-b anchors explanations to the actual prescription, most effectively when reason and action are related.

These findings are accompanied by four limitations. First, weak-label agreement measures instruction-following fidelity, not causal correctness. The taxonomy is a researcher-defined heuristic over observable profile patterns, not validated against expert-annotated churn causes, so agreement is not evidence that the labels identify the true cause of churn. Expert annotation or observed intervention outcomes would give stronger ground truth but were unavailable here.

Second, the reported risk reductions are model-predicted changes, not causal or off-policy validation of real intervention effects---a limitation shared with the algorithmic recourse literature, where recourse without the true structural causal model remains open \cite{karimi2020algorithmic}.

Third, the six-action space does not cover all reason types (no actions target \texttt{LOW\_ENGAGEMENT} or \texttt{OTHER}), and its transformations---detailed in Appendix~\ref{app:actions}---remain study-specific design choices rather than validated causal interventions. Future work should expand and externally validate the action vocabulary \cite{verma2024counterfactual}; the scaling test (Appendix~\ref{app:scaling}) shows such expansion needs hierarchical or diversity-aware retrieval rather than flat top-$k$.

Fourth, $\lambda{=}0.25$ and $k{=}5$ are empirical defaults chosen on the same data used for evaluation, without a separate validation split, and all reported results were obtained on a single dataset. Table~\ref{tab:exp4} shows model-predicted risk reduction is flat within the risk-first regime but collapses by 59\% once $\lambda$ passes 0.50, so the qualitative conclusion is robust while the boundary location is specific to the action costs and feature transformations used in this study. Cross-dataset robustness of both the $\lambda$ boundary and the CF advantage remains future work.

%% ─────────────────────────────────────────────────────────────────────
\section{Conclusion}
\label{sec:conclusion}
%% ─────────────────────────────────────────────────────────────────────

We proposed CARRE, a three-stage pipeline integrating RAG-based retrieval, cost-aware CF scoring, and LLM reasoning for churn prescription. On the IBM Telco dataset, CARRE substantially outperforms SHAP-based baselines in model-predicted risk reduction, and the advantage persists under the cost-controlled SHAP+Cost comparison, indicating it stems from direct CF simulation rather than cost weighting. On the 136-case reason-stratified sample, diagnosis-driven prompt refinement raised weak-label agreement from 79.4\% to 90.4\%; we treat this as a prompt-development result, not an independent generalization estimate. For the 135 v2 explanations, two cross-vendor judges assign mean scores of 4.02--5.00, and a deterministic audit finds no contradictions among 66 verifiable profile claims. Its modular design---retrieval, optimization, LLM justification---lets each component be evaluated and replaced independently.

Future work includes expanding the action space for \texttt{LOW\_\allowbreak ENGAGEMENT} and \texttt{OTHER}, improving faithfulness via citation-based generation \cite{gao2023enabling}, and extending CARRE to other subscription industries.

%% ─────────────────────────────────────────────────────────────────────
\begin{acks}
This research was supported by the Korea Institute for Advancement of Technology (KIAT) grant funded by the Korean Government (MOTIE) (RS-2024-00416131, HRD Program for Industrial Innovation).
\end{acks}

%% ─────────────────────────────────────────────────────────────────────
\bibliographystyle{ACM-Reference-Format}

\clearpage
%% ─────────────────────────────────────────────────────────────────────
\appendix
%% ─────────────────────────────────────────────────────────────────────

These appendices provide the implementation and diagnostic details that complement the main results. They document the exact action-to-feature transformations and eligibility rules, the Stage~3-a prompt template, the base-classifier comparison for counterfactual scoring, the action-space scaling stress test, the per-reason weak-label agreement breakdown, the baseline prescription mappings used in the main experiments, and the implementation and reproducibility details for all reported experiments.

\section{Action-to-Feature Transformations and Eligibility Rules}
\label{app:actions}

Table~\ref{tab:transforms} specifies, for each action, the eligibility condition checked in Stage~2, the exact feature transformation applied to construct the counterfactual profile, and the researcher-defined relative cost, taken directly from the study code. \texttt{A\_CONTRACT\_24M} changes only the \texttt{Contract} field and applies no monthly-charge discount; only \texttt{A\_PRICE\_DISCOUNT} reduces charges, by 8\%. Eligibility is enforced in Stage~2, which excludes ineligible actions before scoring and additionally restricts \texttt{A\_PRICE\_DISCOUNT} to customers with \texttt{MonthlyCharges}${\geq}80$.

\begin{table}[H]
\centering
\caption{Action-to-feature transformations, eligibility conditions, and relative costs, as implemented in the study code (\texttt{apply\_action}, \texttt{eligible}).}
\label{tab:transforms}
\scriptsize
\setlength{\tabcolsep}{3pt}
\begin{adjustbox}{max width=\linewidth}
\begin{tabular}{lp{2.4cm}p{3.2cm}c}
\toprule
Action ID & Eligibility & Feature transformation & Cost \\
\midrule
\texttt{A\_PRICE\_DISCOUNT} & \texttt{MonthlyCharges}${\geq}80$ & \texttt{MonthlyCharges}, \texttt{TotalCharges} ${\times}\,0.92$ & 3.0 \\
\texttt{A\_CONTRACT\_12M} & \texttt{Contract}=Month-to-month & \texttt{Contract}${:=}$One year & 2.0 \\
\texttt{A\_CONTRACT\_24M} & \texttt{Contract}$\in$\{M-t-M, One year\} & \texttt{Contract}${:=}$Two year (no discount) & 3.5 \\
\texttt{A\_PAYMENT\_AUTOPAY} & \texttt{PaymentMethod}=Electronic check & \texttt{PaymentMethod}${:=}$Bank transfer (auto.); \texttt{PaperlessBilling}${:=}$Yes & 1.0 \\
\texttt{A\_ADD\_ONLINESEC} & \texttt{InternetService}$\neq$No, \texttt{OnlineSecurity}=No & \texttt{OnlineSecurity}${:=}$Yes & 2.0 \\
\texttt{A\_ADD\_TECHSUPPORT} & \texttt{InternetService}$\neq$No, \texttt{TechSupport}=No & \texttt{TechSupport}${:=}$Yes & 2.0 \\
\bottomrule
\end{tabular}
\end{adjustbox}
\end{table}

\section{Stage 3-a Prompt Template}
\label{app:prompt}

This appendix gives a condensed schematic of the Stage~3-a prompt. The actual prompt uses a fixed plain-text schema: the model returns a \texttt{Reason:} line, a \texttt{Plan:} line, and two rationale bullets, which are parsed downstream. The version shown is the \emph{v2} scaled-evaluation prompt used for the 79.4\% diagnostic result, the explanations evaluated in Table~\ref{tab:judge}, and the 136-case retrieval ablation in Table~\ref{tab:exp7}. Table~\ref{tab:exp5} used an earlier model-comparison formulation. The \emph{v3} prompt that yields 90.4\% adds an enforcement block comprising lowest-step-wins tie-breaking with worked examples and an explicit all-three-conditions requirement for Step~4; it is otherwise identical to the v2 scaled-evaluation prompt.

The design encodes three choices, each addressing a failure observed during development. First, the retrieved \texttt{cf\_action} is deliberately withheld, so the model reasons forward from the customer profile to a reason label rather than rationalizing backward from an action it has already been shown. Second, the six classification rules are listed explicitly and in priority order, turning reason assignment into a deterministic rule-application task. Third, the \emph{CoT Instruction} block asks the model to check each rule in turn and state the first one it satisfies before committing to a label; our refinement experiments found this the single most effective ingredient for correcting step-order errors. Finally, the \texttt{Plan} output is instructed to select from the retrieved action set; out-of-set action IDs are counted as auxiliary-plan violations.

\begin{lstlisting}
[System]
You are a churn retention analyst. Given a customer profile
and retrieved intervention documents, classify the primary
churn reason and recommend one action.

[Input]
Customer: {"Contract":"Month-to-month","tenure":3,
           "MonthlyCharges":85.5,"PaymentMethod":
           "Electronic check","TechSupport":"No",...}
Churn probability: f(x) = 0.82
Retrieved actions: [A_PRICE_DISCOUNT, A_CONTRACT_24M, ...]
(cf_action is NOT provided)

[Classification Rules -- evaluate in priority order]
Step 1: MonthlyCharges>=80 AND Month-to-month
                                      -> PRICE_SENSITIVE
Step 2: Month-to-month AND tenure<=6  -> CONTRACT_RISK
Step 3: Electronic check payment      -> PAYMENT_FRICTION
Step 4: InternetService!=No AND TechSupport=No
        AND OnlineSecurity=No          -> SUPPORT_DEFICIT
Step 5: tenure<=3                      -> LOW_ENGAGEMENT
Step 6: (otherwise)                   -> OTHER

[CoT Instruction]
Evaluate each step in order. State which condition is first
satisfied. Assign that label. Do not infer reason from
the recommended action.

[Output -- fixed plain-text schema]
Reason: <LABEL>
Plan: <ACTION_ID from retrieved set only or NONE>
- <profile-grounded rationale bullet 1>
- <profile-grounded rationale bullet 2>
\end{lstlisting}

\section{Base Classifier Comparison for CF Scoring}
\label{app:model}

We compare four calibrated base models---LR with sigmoid calibration and three tree ensembles with isotonic calibration (gradient boosting, random forest, histogram gradient boosting)---on predictive quality and CF-surface smoothness. Smoothness is measured by sweeping a graded price discount (0--30\% in 31 steps) on the 313 high-risk cases and recording how the predicted churn probability responds. \emph{Distinct} is the fraction of distinct probability values along the sweep (1.0 = fully continuous). \emph{Flat} is the fraction of adjacent steps with $|\Delta p| < 10^{-4}$, i.e., dead zones where a small feature perturbation registers no change in predicted risk. \emph{Max jump} is the mean largest single-step $|\Delta p|$, capturing discontinuous cliffs. Table~\ref{tab:model_cf} below reports the results. Gradient boosting attains marginally higher ROC-AUC of 0.845 against 0.842 and PR-AUC of 0.659 against 0.633, confirming a marginal ranking advantage for gradient boosting. All three tree models nevertheless produce coarse, discontinuous CF surfaces, whereas logistic regression yields the fully continuous, graded response that Stage~2 requires.

The practical takeaway is that counterfactual scoring rewards a smooth probability surface more than raw ranking power. A simple, well-calibrated logistic regression therefore serves CARRE better than a stronger tree ensemble: it keeps predicted-risk changes estimable, it keeps the pipeline easy to interpret and deploy, and it gives up almost nothing in ranking accuracy. For these reasons we adopt calibrated logistic regression as CARRE's base estimator.

\begin{table}[H]
\centering
\caption{Base classifier comparison for CF scoring: predictive quality vs.\ CF-surface smoothness (313 high-risk cases; graded 0--30\% discount sweep, 31 steps).}
\label{tab:model_cf}
\small
\begin{adjustbox}{max width=\linewidth}
\begin{tabular}{lcccccc}
\toprule
Base model & ROC-AUC & PR-AUC & Brier & Distinct$\uparrow$ & Flat$\downarrow$ & Max jump$\downarrow$ \\
\midrule
\textbf{LR + sigmoid (CARRE)} & 0.842 & 0.633 & 0.138 & \textbf{1.00} & \textbf{0\%} & \textbf{0.002} \\
GBM + isotonic      & 0.845 & 0.659 & 0.136 & 0.30 & 64\% & 0.087 \\
RF + isotonic       & 0.831 & 0.629 & 0.141 & 0.29 & 66\% & 0.059 \\
HistGBM + isotonic  & 0.834 & 0.639 & 0.140 & 0.48 & 45\% & 0.098 \\
\bottomrule
\end{tabular}
\end{adjustbox}
\end{table}

\section{Action-Space Scaling Stress Test}
\label{app:scaling}

This is a retrieval-only stress test: the six action documents were augmented with up to 394 synthetic lever documents (retrieval documents, not executable actions)---parameter variants of contract terms, discount tiers, segment-targeted offers, and channel campaigns, written in the same document style as the real actions---and the downstream CF and LLM stages were not run. For each library size we measure retrieval runtime and the survival of the CF-oracle action in the top-$k$ over all 313 high-risk cases (Table~\ref{tab:scaling}). Query encoding ($\approx$1\,ms) dominates end-to-end latency regardless of library size, and index build stays under 7 seconds. Exact-action Recall@5 collapses as soon as near-duplicate parameter variants are added ($|\mathcal{A}|{\geq}12$), because inner-product ranking cannot distinguish among semantically near-identical variants of the same lever. Family-level recall---whether \emph{any} document of the oracle action's lever family survives in the top-5---remains high (0.92--1.00) up to $|\mathcal{A}|{=}100$ but collapses beyond $|\mathcal{A}|{\approx}200$, once a single high-similarity lever family grows large enough to monopolize the entire top-$k$. Hierarchical retrieval (family-level candidate grouping followed by within-family CF simulation) or diversity-aware retrieval such as maximal marginal relevance addresses this failure mode.

\begin{table}[H]
\centering
\captionsetup{skip=1pt}
\caption{Action-space scaling stress test (313 high-risk cases; synthetic levers styled after the real action documents). Query latency is encoding + search per query. Exact = CF-oracle action ID retrieved; Family = any document of the oracle's lever family retrieved.}
\label{tab:scaling}
\scriptsize
\setlength{\tabcolsep}{2.5pt}
\renewcommand{\arraystretch}{0.86}
\begin{adjustbox}{max width=\linewidth}
\begin{tabular}{rccccc}
\toprule
$|\mathcal{A}|$ & Query (ms) & Exact R@5 & Exact R@20 & Family R@5 & Family R@10 \\
\midrule
6   & 1.40 & 0.853 & 1.000 & 1.000 & 1.000 \\
12  & 1.23 & 0.000 & 1.000 & 1.000 & 1.000 \\
25  & 0.81 & 0.000 & 0.703 & 0.997 & 1.000 \\
50  & 1.23 & 0.000 & 0.000 & 0.987 & 1.000 \\
100 & 1.09 & 0.000 & 0.000 & 0.920 & 0.987 \\
200 & 0.97 & 0.000 & 0.000 & 0.010 & 0.920 \\
400 & 1.23 & 0.000 & 0.000 & 0.003 & 0.010 \\
\bottomrule
\end{tabular}
\end{adjustbox}
\end{table}

\section{Per-Reason Weak-Label Agreement}
\label{app:reason}

Table~\ref{tab:exp5_reason} breaks down the four-model comparison of Section~\ref{sec:exp_llm} by reason category. The breakdown makes clear that the aggregate agreement gap is not a uniform accuracy difference but is concentrated in a few categories. GPT-4o classifies almost every category correctly, whereas the weaker models fail selectively. GPT-4o-mini collapses on the price and contract categories, and both Llama-3.3-70B and Qwen3-32B score zero on \texttt{CONTRACT\_RISK} and \texttt{OTHER} while still handling simpler categories such as \texttt{PRICE\_SENSITIVE}. This selective failure, rather than an even drop across all categories, is what separates the models, and it suggests that the smaller models fall back on a few surface cues instead of applying the full priority order.

\begin{table}[H]
\centering
\caption{Per-reason weak-label agreement by model (30-case set, 5 cases per reason).}
\label{tab:exp5_reason}
\small
\begin{adjustbox}{max width=\linewidth}
\begin{tabular}{lcccc}
\toprule
Reason & GPT-4o & GPT-4o-mini & Llama-70B & Qwen3-32B \\
\midrule
PRICE\_SENSITIVE  & 5/5~(100\%) & 0/5~(0\%)   & 5/5~(100\%) & 4/5~(80\%) \\
CONTRACT\_RISK    & 4/5~(80\%)  & 0/5~(0\%)   & 0/5~(0\%)   & 0/5~(0\%) \\
PAYMENT\_FRICTION & 5/5~(100\%) & 5/5~(100\%) & 3/5~(60\%)  & 0/5~(0\%) \\
SUPPORT\_DEFICIT  & 5/5~(100\%) & 1/5~(20\%)  & 3/5~(60\%)  & 2/5~(40\%) \\
LOW\_ENGAGEMENT   & 5/5~(100\%) & 2/5~(40\%)  & 4/5~(80\%)  & 5/5~(100\%) \\
OTHER             & 4/5~(80\%)  & 4/5~(80\%)  & 0/5~(0\%)   & 0/5~(0\%) \\
\midrule
Overall           & 0.933 & 0.400 & 0.500 & 0.367 \\
\bottomrule
\end{tabular}
\end{adjustbox}
\end{table}

\section{Baseline Prescription Mappings}
\label{app:baselines}

Table~\ref{tab:baselines} gives the rule-based and SHAP-based baseline mappings, taken directly from the study code. The rule-based baseline applies the conditions in priority order and returns \texttt{NONE} when none holds. The SHAP baseline maps the feature with the highest positive churn-directed SHAP value to a candidate action; ties and ineligible targets fall through a fixed priority list. The random baseline draws one eligible action per customer under a fixed per-customer seed (no averaging over repetitions).

\begin{table}[H]
\centering
\caption{Rule-based (profile condition $\to$ action, in priority order) and SHAP-based (feature $\to$ action) baseline mappings, as implemented in the study code.}
\label{tab:baselines}
\scriptsize
\setlength{\tabcolsep}{3pt}
\begin{adjustbox}{max width=\linewidth}
\begin{tabular}{ll}
\toprule
Rule-based condition & Action \\
\midrule
\texttt{MonthlyCharges}${\geq}80$ \& Month-to-month & \texttt{A\_PRICE\_DISCOUNT} \\
Month-to-month & \texttt{A\_CONTRACT\_12M} \\
\texttt{PaymentMethod}=Electronic check & \texttt{A\_PAYMENT\_AUTOPAY} \\
\texttt{InternetService}$\neq$No \& \texttt{TechSupport}=No & \texttt{A\_ADD\_TECHSUPPORT} \\
\texttt{InternetService}$\neq$No \& \texttt{OnlineSecurity}=No & \texttt{A\_ADD\_ONLINESEC} \\
otherwise & \texttt{NONE} \\
\midrule
SHAP top feature & Action \\
\midrule
\texttt{MonthlyCharges}, \texttt{TotalCharges} & \texttt{A\_PRICE\_DISCOUNT} \\
\texttt{Contract}, \texttt{tenure} & \texttt{A\_CONTRACT\_12M} \\
\texttt{PaymentMethod} & \texttt{A\_PAYMENT\_AUTOPAY} \\
\texttt{TechSupport} & \texttt{A\_ADD\_TECHSUPPORT} \\
\texttt{OnlineSecurity} & \texttt{A\_ADD\_ONLINESEC} \\
\bottomrule
\end{tabular}
\end{adjustbox}
\end{table}

\section{Implementation and Reproducibility Details}
\label{app:repro}

All experiments use seed~42 and an 80/20 stratified split (5{,}634 train / 1{,}409 test). After dropping \texttt{customerID} and the \texttt{Churn} label, 19 predictors remain. The four numeric features (\texttt{SeniorCitizen}, \texttt{tenure}, \texttt{MonthlyCharges}, \texttt{TotalCharges}) are median-imputed and standardized; categorical features are most-frequent-imputed and one-hot encoded (unknown categories ignored); all preprocessing is fit on the training split only. The churn estimator is a \texttt{CalibratedClassifierCV} wrapping \texttt{LogisticRegression} ($C{=}1.0$, \texttt{max\_iter}${=}2000$, class-balanced) with sigmoid calibration under 3-fold cross-validation. Bootstrap confidence intervals use 2{,}000 per-method (marginal) resamples at seed~42. Retrieval uses \texttt{all-MiniLM-L6-v2} embeddings, L2-normalized, in a FAISS \texttt{IndexFlatIP} (cosine similarity); the scaling stress test generates its synthetic lever documents deterministically at seed~42. LLM calls use \texttt{gpt-4o}, \texttt{gpt-4o-mini}, \texttt{llama-3.3-70b-versatile}, and \texttt{qwen/qwen3-32b} at temperature~0, with the provider-default top-$p$, no API seed, a 400-token cap, and up to three API retries on transient errors; malformed outputs are reparsed by a fixed rule and, on repeated failure, counted as a violation. We used provider aliases rather than pinned model snapshots, which limits exact reproducibility of the LLM evaluations.

\end{document}